\documentclass{article}

\usepackage{microtype}
\usepackage{graphicx}
\usepackage{subfigure}
\usepackage{booktabs} 

\usepackage{hyperref}

\usepackage{amsmath,amsfonts,bm}

\def\eqref#1{equation~\ref{#1}}

\def\1{\bm{1}}

\DeclareMathAlphabet{\mathsfit}{\encodingdefault}{\sfdefault}{m}{sl}
\SetMathAlphabet{\mathsfit}{bold}{\encodingdefault}{\sfdefault}{bx}{n}

\DeclareMathOperator*{\argmax}{argmax}
\DeclareMathOperator*{\argmin}{argmin}

\usepackage{hyperref}
\usepackage{url}
\usepackage{ulem}

\newcommand{\model}{IBO\xspace}

\usepackage{multicol}

\usepackage{amsmath,amsthm,mathtools}
\usepackage{xspace}
\usepackage{enumitem}
\usepackage{microtype}  
\usepackage{multirow}
\usepackage{algorithm}

\usepackage{dblfloatfix}
\usepackage{booktabs}
\usepackage{xcolor}
\usepackage{float}

\theoremstyle{definition}
\newtheorem{rmk}{Remark}

\newcommand{\ie}{\textit{i.e.\xspace}}
\newcommand{\eg}{\textit{e.g.\xspace}}

\newcommand{\expect}{\mathbb E}
\renewcommand{\Pr}{\mathbb P}
\newcommand{\ignore}[1]{}
\renewcommand{\eqref}[1]{Eq.~(\ref{#1})}

\renewcommand{\emph}[1]{\textit{#1}}

\definecolor{darkblue}{rgb}{0.0,0.0,0.55}
\hypersetup{
  colorlinks = true,
  citecolor  = darkblue,
  linkcolor  = darkblue,
  citecolor  = darkblue,
  filecolor  = darkblue,
  urlcolor   = darkblue,
}

\usepackage[accepted]{arxiv2020}

\arxivtitlerunning{Weighting Is Worth the Wait: Bayesian Optimization with Importance Sampling}

\begin{document}

\twocolumn[

\arxivtitle{Weighting Is Worth the Wait:\\
Bayesian Optimization with Importance Sampling}  


\arxivsetsymbol{equal}{*}
\begin{arxivauthorlist}
\arxivauthor{Setareh Ariafar}{ece}
\arxivauthor{Zelda Mariet}{goo}
\arxivauthor{Ehsan Elhamifar}{cs}
\arxivauthor{Dana Brooks}{ece}
\arxivauthor{Jennifer Dy}{ece}
\arxivauthor{Jasper Snoek}{goo}
\end{arxivauthorlist}

\arxivaffiliation{ece}{Department of Electrical and Computer Engineering, Northeastern University}
\arxivaffiliation{goo}{Google Research}
\arxivaffiliation{cs} {Khoury College of Computer Sciences, Northeastern University}

\arxivcorrespondingauthor{Setareh Ariafar}{sariafar@ece.neu.edu}

\arxivkeywords{Bayesian optimization, Gaussian Processes}

\vskip 0.3in
]



\printAffiliationsAndNotice{\arxivEqualContribution} 

\begin{abstract}
    Many contemporary machine learning models require extensive tuning of hyperparameters to perform well. A variety of methods, such as Bayesian optimization, have been developed to automate and expedite this process. However, tuning remains extremely costly as it typically requires repeatedly fully training models. We propose to accelerate the Bayesian optimization approach to hyperparameter tuning for neural networks by taking into account the relative amount of information contributed by each training example. To do so, we leverage importance sampling (IS); this significantly increases the quality of the black-box function evaluations, but also their runtime, and so must be done carefully. Casting hyperparameter search as a multi-task Bayesian optimization problem over both hyperparameters and importance sampling design achieves the best of both worlds: by learning a parameterization of IS that trades-off evaluation complexity and quality, we improve upon Bayesian optimization state-of-the-art runtime and final validation error across a variety of datasets and complex neural architectures.
    \vspace{-2em}
  
\end{abstract}

\section{Introduction}
The incorporation of more parameters and more data, coupled with faster computing and longer training times, has driven state-of-the-art results across a variety of benchmark tasks in machine learning.  However, careful model tuning remains critical in order to  find good configurations of hyperparameters, architecture and optimization settings. This tuning requires significant experimentation, training many models, and is often guided by expert intuition, grid search, or random sampling. 
Such experimentation multiplies the cost of training, and incurs significant financial, computational, and even environmental costs~\citep{strubell19}.

Bayesian optimization (BO) offers an efficient alternative when the tuning objective can be effectively modeled by a surrogate regression~\citep{bergstra2011algorithms,snoek2012practical}, or when one can take advantage of related tasks~\citep{swersky2013multi} or strong priors over problem structure~\citep{swersky2014freeze, domhan2015speeding}. BO optimizes an expensive function by iteratively building a relatively cheap probabilistic surrogate and evaluating a carefully balanced combination of uncertain and promising regions (exploration {\it vs.} exploitation). 

In the context of neural network hyperparameter optimization, BO typically involves an inner loop of training a model 
given a hyperparameter configuration, and then evaluating validation error as the objective to be optimized.  This inner loop is expensive and its cost grows with the size of the dataset: querying modern models even once may require training for days or weeks.

One strategy to mitigate the high cost of hyperparameter tuning is to enable the BO algorithm to trade off between the value of the information gained from evaluating a hyperparameter setting and the cost of that evaluation. For example, \citet{swersky2013multi} and \citet{klein2016fast} allow BO to evaluate models trained on randomly chosen subsets of data to obtain more, but less informative, evaluations. We propose an alternative approach: our method, Importance-based Bayesian Optimization (IBO), dynamically learns when spending additional effort training a network to obtain a higher fidelity observation is worth the incurred cost. To achieve this, in addition to considering the hyperparameters, IBO takes into account the underlying training data and focuses the computation on more informative training points. Specifically, IBO models a distribution over the location of the optimal hyperparameter configuration, and allocates experimental budget according to cost-adjusted expected reduction in entropy~\cite{hennig2012entropy}. Therefore, higher fidelity observations provide a greater reduction in entropy, albeit at a higher evaluation cost.

\begin{figure*}[t]
  \includegraphics[width=\textwidth]{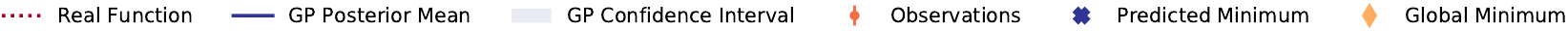}
  \vspace{-2em}  

  \subfigure[GP model of $x \to x\sin x$ with many noisy points]{ \includegraphics[width=0.5\linewidth]{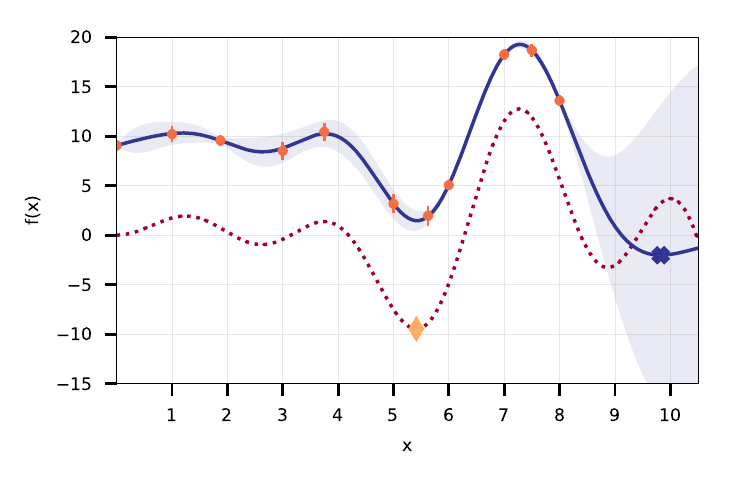}}
  \subfigure[GP model of $x \to x\sin x$ with few noiseless points]{\vspace{-5em} \includegraphics[width=0.5\linewidth]{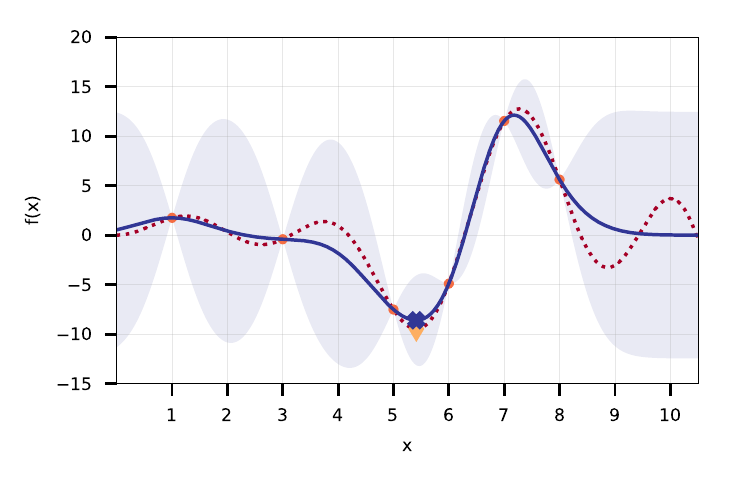}}
  \vspace{-1em}
\caption{Motivating example: compared to querying many points with lower fidelity (left), observing few points with higher fidelity (right) can significantly improve the model's predicted minimum location (here, lower-fidelity queries always \emph{over-estimate} the target function, reflecting the impact of limiting the number of SGD iterations during training). However, obtaining higher-quality estimates can significantly slow down the overall runtime of Bayesian optimization; by learning to optimize the tradeoff between the value and cost of obtaining high-fidelity estimates, we show in Section~\ref{sec:experiments} that IBO achieves the best of both worlds.}
\label{fig:GP-illus}
\vspace{-1em}
\end{figure*}

To decide how much effort to allocate to training a network and which training examples to prioritize, IBO leverages both the properties of stochastic gradient descent (SGD) and recent work on importance sampling~\citep{NIPS2018_7957}. At each SGD iteration, IBO estimates how much each training example will impact the model; based on this estimate, IBO either does a normal round of SGD, or a more costly importance-weighted gradient update.  

Balancing the cost of the inner loop of neural network training and the outer loop of BO is a non-trivial task; if done naively, the overall hyperparameter tuning procedure will be substantially slower. To address this issue, we adopt a multi-task Bayesian optimization formulation for IBO and develop an algorithm that dynamically adjusts for the trade-off between the cost of training a network at higher fidelity and getting more but noisier evaluations (Fig. \ref{fig:GP-illus}).
This approach allows us to obtain higher quality black-box function evaluations only when worthwhile, while controlling the average cost of black-box queries. As a consequence, we are able to tune complex network architectures over challenging datasets in less time and with better results than existing state-of-the-art BO methods. Tuning a ResNet on CIFAR-100, IBO improves the validation error by $\ge 4\%$ over the next best method; other baselines are not able to reach IBO's performance, even with additional computational budget.

\vspace{-.5em}
\paragraph{Contributions.} We introduce a multi-task Bayesian optimization framework, \model (\textbf{I}mportance-based \textbf{B}ayesian \textbf{O}ptimization), which takes into account the contribution of each training point during the evaluation of a candidate hyperparameter. To do so, \model optimizes the importance sampling tradeoff between quality and runtime while simultaneously searching hyperparameter space. We show on extensive benchmark experiments that the computational burden incurred by importance sampling is more than compensated for by the principled search through hyperparameter space that  it enables. We show across these experiments that \model consistently improves over a variety of baseline Bayesian optimization methods. On more complex datasets, \model converges significantly faster in wall-clock time than existing methods and furthermore reaches lower validation errors, even as other methods are given larger time budgets. 
\section{Related work}
Several different methods have been proposed to accelerate the hyperparameter tuning process. \citet{swersky2013multi} proposed Multi-Task Bayesian Optimization (MTBO), which performs surrogate cheap function evaluations on a small subset of training data which is then used to extrapolate the performance on the entire training set. Motivated by this work, \citet{klein2016fast} introduced Fabolas, which extends MTBO to also learn the sufficient size of training data. MTBO and Fabolas avoid costly function evaluations by training on small datasets where data is uniformly chosen at the beginning of each training round. 

Another body of related work involves modeling the neural network's loss as a function of both the hyperparameters and the inner training iterations. Then, the goal is to extrapolate and forecast the ultimate objective value and stop underperforming training runs early. Work such as \citep{swersky2014freeze,domhan2015speeding,dai2019bayesian,golovin17} falls under  this category. These methods generally have to deal with the cubic cost of Gaussian processes --- $O(n^3t^3)$, for $n$ observed hyperparameters and $t$ iterations. In practice, these methods typically apply some type of relaxation. For example, the freeze-thaw method~\citet{swersky2014freeze} assumes that  training curves for different hyperparameter configurations are independent conditioned on their prior mean, which is drawn from another global GP.

Moreover, an alternative approach to Bayesian optimization solves the hyperparameter tuning problem through enhanced random search. Hyperband~\citep{li2016hyperband} starts from several randomly chosen hyperparameters and trains them on a small subset of data. Following a fixed schedule, the algorithm stops underperforming experiments and then retrains the remaining ones on larger training sets. Hyperband outperforms standard BO in some settings, as it is easily parallelized and not subject to model misspecification. However, Hyperband's exploration is necessarily limited to the initial hyperparameter sampling phase: the best settings chosen by Hyperband inevitably will correspond to one of initial initializations, which were selected uniformly and in an unguided manner. To address this issue, several papers, including \citep{falkner2018bohb,wang2018combination,bertrand2017hyperparameter}, have proposed the use of Bayesian optimization to warm-start Hyperband and perform a guided search during the initial hyperparameter sampling phase.

Finally, IBO belongs to the family of multi-fidelity Bayesian optimization methods~\citep{kandasamy2016gaussian,soton64698,Huang2006SequentialKO,klein2016fast} methods, which take advantage of cheap approximations to the target black-box function. Of those methods, Fabolas~\citep{klein2016fast} focuses specifically on hyperparameter tuning, and is included as a baseline in all our experiments. Fabolas uses cheap evaluations of the network validation loss by training the network on a randomly sampled subset of the training dataset. Hence, both IBO and Fabolas depend directly on training examples to vary the cost of querying the black-box function; Fabolas by using fewer examples for cheap evaluations, whereas IBO uses the per-example contribution to training to switch to \emph{costlier} evaluations. 

Existing literature on hyperparameter tuning weighs all training examples equally and does not take advantage of their decidedly unequal influence. To the best of our knowledge, IBO is the first method to exploit the informativeness of training data to accelerate hyperparameter tuning, merging Bayesian optimization with importance sampling. 

\vspace{-1mm}
\paragraph{Terminology.}
We refer to one stochastic gradient descent (SGD) update to a neural network as an \emph{inner optimization round}. Conversely, an \emph{outer optimization round} designates one iteration of the BO process: fitting a GP, optimizing an acquisition function, and evaluating the black-box function.
\section{Importance sampling for BO}
Bayesian optimization is a strategy for the global optimization of a potentially noisy, and generally non-convex, black-box function $f: \mathcal X \to \mathbb R$.  The function $f$ is presumed to be expensive to evaluate in terms of time, resources, or both. 

In the context of hyperparameter tuning, $\mathcal X$ is the space of hyperparameters, and $f(x)$ is the validation error of a neural network trained with hyperparameters $x$. 

Given a set $\mathcal D = \{(x_i, y_i = f(x_i))\}_{i=1}^N$ of hyperparameter configurations $x_i$ and associated function evaluations $y_i$ (which may be subject to observation noise), Bayesian optimization starts by building a surrogate model for $f$ over $\mathcal D$. Gaussian processes (GPs), which provide a flexible non-parametric distribution over smooth functions, are a popular choice for this probabilistic model, as they provide tractable closed-form inference and facilitate the specification of a prior over the functional form of  $f$~\citep{rasmussen2003gaussian}. 

\subsection{Surrogate model quality vs. computational budget}
Given a zero-mean prior with covariance function $k$, the GP's posterior belief about the unobserved output $f(x)$ at a new point $x$ after seeing data $\mathcal D = \{x_i, y_i\}_{i=1}^N$ is a Gaussian distribution with mean $\mu(x)$ and variance $\sigma^2(x)$ such that
\begin{equation} 
    \label{eq:GP-post-mu}
    \begin{aligned}
        \mu&(x) = \mathbf{k}(x)^T \Big(\mathbf{K} + \sigma_{\text{noise}}^2 I\Big)^{-1} y, \\ 
        \sigma^2&(x) = k(x,x) - \mathbf{k}(x)^T\Big(\mathbf{K} + \sigma_{\text{noise}}^2 I\Big)^{-1}\mathbf{k}(x),
    \end{aligned}
\end{equation}
where $\mathbf{k}(x) = [k(x,x_i)]_{i=1}^N$, $\mathbf{K} = [k(x_i,x_j)]_{i,j=1}^N$, and $\sigma_{\text{noise}}^2$ is the variance of the observation noise, that is, $y_i \sim \mathcal N(f(x_i), \sigma_{\textup{noise}}^2)$.

Given this posterior belief over the value of unobserved points, Bayesian optimization selects the next point (hyperparameter set) $x$ to query by solving
\begin{equation}
  \label{eq:BO-af}
  x = \argmax_{x \in \mathcal X} \alpha(x \mid \mathcal D),
\end{equation}
where $\alpha( \cdot)$ is the \emph{acquisition function}, which quantifies the expected added value of querying $f$ at point $x$, based on the posterior belief on $f(x)$ given by \eqref{eq:GP-post-mu}. 

Typical choices for the acquisition function $\alpha$ include entropy search (ES) \citep{hennig2012entropy} and its approximation predictive entropy search~\citep{hernandez2014predictive}, knowledge gradient \citep{wu2017bayesian}, expected improvement~\citep{movckus1975bayesian,jones1998efficient} and upper/lower confidence bound~\citep{cox1992statistical,Cox97sdo:a}. 

Entropy search quantifies how much knowing $f(x)$ reduces the entropy of the distribution $\Pr[x^* \mid \mathcal D]$ over the location of the best hyperparameters $x^*$:
\begin{align}
    \alpha_{\textup{ES}}(x \mid D) = \expect_{y \mid x, \mathcal D} \Big[H&\big(\Pr [x^* \mid \mathcal D]\big) \label{eq:es}\\
    &- H\big(\Pr[x^* \mid \mathcal D \cup \{x, y\}]\big)\Big], \nonumber
\end{align}
where $H$ is the entropy function and the expectation is taken with respect to the posterior distribution over the observation $y$ at hyperparameter $x$.

The more accurate the observed values $y$, the more accurate the GP surrogate model (\ref{eq:GP-post-mu}). A more accurate surrogate model, in turn, defines a better acquisition function (\ref{eq:BO-af}), and, finally, a more valuable Bayesian optimization outer loop. 

Previous work has tackled this trade-off during the BO process by early-stopping training that is predicted to yield poor final values~\citep{swersky2014freeze,dai2019bayesian}. IBO takes the opposite route, detecting when to spend \emph{additional} effort to acquire a more accurate value of $f(x)$.

Crucially, hyperparameter tuning for neural networks is not an entirely black-box optimization setting, as we know the loss minimization framework in which neural networks are trained. We take advantage of this by allocating computational budget \emph{at each SGD iteration}; based on the considered training points, IBO switches from standard SGD updates to the more computationally intensive importance sampling updates. This is the focus of the following section.

\subsection{Importance sampling for loss minimization}
The impact of training data points on one (batched) SGD iteration has benefited from significant attention in machine learning~\cite{NIPS2014_5355,pmlr-v38-schmidt15,Zhang2017DeterminantalPP,pmlr-v54-fu17a}. For the purposes of IBO, we focus on importance sampling (IS)~\citep{NIPS2014_5355,pmlr-v37-zhaoa15}. IS minimizes the variance in SGD updates;\footnote{IS also benefits SGD with momentum~\citep{NIPS2018_7957}. Although we focus our analysis on pure SGD, IBO also extends to certain SGD variants.} however, IS is parameterized by the per-example gradient norm for the current weights of the network, and as such incurs a significant computational overhead.

Specifically, let $g(w) = \tfrac 1m\sum_{i=1}^m g_i(w)$ be the training loss, where $m$ is the number of training examples and $g_i$ is the loss at point $i$. To minimize $g(w)$, SGD with importance sampling iteratively computes estimate $w_{t+1}$ of $w^* = \argmin g$ by sampling $i \in \{1, \ldots, m\}$ with probability $p_i \propto \|\nabla g_i(w_t)\|$, then applying the update
\begin{equation}
    \label{eq:is}
    w_{t+1} = w_t - \eta \frac 1{mp_i} \nabla g_i(w_t),
\end{equation}
where $\eta$ is the learning rate. Update (\ref{eq:is}) provably minimizes the variance of the gradient estimate, which in turn improves the convergence speed of SGD.\footnote{Standard SGD is recovered by setting $p_i = 1/m$.} 

Various solutions to efficiently leverage importance sampling have been suggested
~\citep{pmlr-v37-zhaoa15}. We leverage recent work~\citet{katharopoulos2018not}, which speeds up batched SGD  with IS by a cheap subroutine that determines whether IS's variance reduction justifies the incurred computational cost at each SGD step.

To achieve efficient IS for batches of size $b$, \citet{katharopoulos2018not} introduce a \emph{pre-sample batch size} hyperparameter $B \ge b$. At each SGD step, $B$ points are first sampled uniformly at random, from which a batch of size $b$ is then subsampled. These $b$ points are sampled either uniformly or with importance sampling, depending on an upper bound on the variance reduction permitted by IS. 

\subsection{Multi-task BO for importance sampling}
In~\citep{katharopoulos2018not}, the authors state that the added value of importance sampling is extremely sensitive to the number $B$ of the pre-sampled data points; we verify this empirically in \S\ref{sec:experiments}, showing that naively replacing standard SGD with the IS algorithm of~\citep{katharopoulos2018not} does not improve upon standard BO hyperparameter tuning. To maximize the utility of importance sampling, we instead opt for a multi-task BO framework, within which the search through hyperparamater space $\mathcal X$ is done in parallel to a second task: optimization over $B$.

Multi-task Bayesian optimization (MTBO) \citep{swersky2013multi} extends BO to evaluating a point $x$ on multiple correlated tasks. To do so, MTBO optimizes an objective function over a \textit{target task} which, although expensive to evaluate, provides the maximum utility for the downstream task. MTBO exploits cheap evaluations on surrogate tasks to extrapolate performance on the target task; here, the target task evaluates $f(x)$ when sampling a batch from all training data, whereas the surrogate task evaluates $f(x)$ when subsampling from a super-batch of $B$ datapoints at each SGD iteration
MTBO uses the entropy search acquisition function (Eq.~\ref{eq:es}), and models an objective function over points $x \in \mathcal X$ and tasks $t \in \mathcal T$ via a multi-task GP \citep{journel1978mining,bonilla2008multi}. The covariance between two pairs of points and corresponding tasks is defined through a Kronecker product kernel:
\begin{align}
    \label{eq:mt-gp}
    k\big((x,t),(x',t')\big) = k_X(x,x')\cdot k_T(t,t'),
\end{align}
where $k_X$ models the relation between the hyperparameters and $k_T$ describes the correlation between tasks. 

For our case, the subsampling size $B$ is the task variable while the optimal task sets $B^*$ to the size of the entire training set. Let $f_n(x_i\mid B_i)$ denote the validation error value at hyperparameter $x_i$ after $n$ training iterations using IS with pre-sample size $B_i$. We define the multi-task kernel for the GP that models $f_n(x_i \mid B_i)$ as \[k^{(f)}\big((x, B), (x', B')\big) = k^{(f)}_X(x,x') \cdot k^{(f)}_B(B,B'),\]with the sub-task kernels defined as
\begin{equation}
\label{eq:task-kernel}
\begin{aligned}
    k^{(f)}_X(x, x') &= \text{Mat{\'e}rn}_{5/2}(x,x')\\
    k^{(f)}_B(B, B') &= (1-B)^\nu(1-B')^\nu + 1.
\end{aligned}
\end{equation}
Additionally, following~\cite{snoek2012practical}, we penalize the evaluation of any point $(x, B)$ by the computational cost $c_n(x \mid B)$ of training a model for $n$ SGD iterations at hyperparameter $x$ with subsampling size $B$. This penalty guides the hyperparameter search towards promising yet relatively inexpensive solutions. We model the training cost $c_n(x \mid B)$ using a multi-task GP fitted to the log cost of observations $c_{i|n}$ that are collected during BO. We choose the covariance function \[k^{(c)}\big((x, B), (x', B')\big) = k^{(c)}_X(x,x') \cdot k^{(c)}_B(B,B'),\]  where this time we modify the kernel on $B$ to reflect that larger $B$ increases training time:
\begin{equation}
\label{eq:cost-kernel}
\begin{aligned}
    k^{(c)}_X(x, x') &= \text{Mat{\'e}rn}_{5/2}(x,x'),\\
    k^{(c)}_B(B, B') &= B^\lambda B'^\lambda + 1.
\end{aligned}
\end{equation}
Our choices for $k^{(f)}$ and $k^{(c)}$ follow \citep{klein2016fast}, who recommend the associated feature maps. 

Our resulting acquisition function is thus:
\begin{align} \label{eq:IBO-af}
    \alpha_n(x, B) =& \frac 1 { \mu(c_n(x \mid B))}\Big[H(\Pr[x^* \mid B^*,\mathcal D_n]) \\ \nonumber 
    &- \expect_{y} \big[ H(\Pr[x^* \mid B^*, \mathcal D_n \cup \{x, B, y\}]) \big] \Big], \label{eq:BO-IS-B-af}
\end{align}
where $ \mu(c_n(x \mid B))$ is the posterior mean of the GP modeling the training cost; as previously,  $\Pr(x^* \mid B^*,\mathcal D_n)$ is the probability that $x^*$ is the optimal solution at the target task  $B^*$ given data $\mathcal D_n= \{x_i, B_i, y_{i|n}, c_{i|n}\}_{i=1}^N$. 

Our algorithm is presented in Algorithm~\ref{alg:IBO}. The initialization phase follows the MTBO convention: we collect initial data at randomly chosen inputs $x$, and evaluate each hyperparameter configuration with a randomly selected value for $B$. \texttt{DoSGD} is the subroutine proposed by~\citep{katharopoulos2018not}; it determines if the variance reduction enabled by importance sampling is worth the additional cost at the current SGD iteration.

\begin{algorithm}[tb]
   \caption{Importance-based BO}\label{alg:IBO}
\begin{algorithmic}
   \STATE{Obtain initial data $\mathcal D_n = \{x_i, B_i, y_{i|n}, c_{i|n}\}$}
\FOR{$i=1,\dots, n_\textup{BO}$}
    \STATE {Fit multi-task GPs to $f_n$ and $c_n$ given $\mathcal D_n$}
    \STATE{$x,B \leftarrow \argmax \alpha_n(x,B \mid \mathcal D_n)$}
    \STATE{$\mathcal M \leftarrow$ model initialized with hyperparams $x$}
        \FOR{$j=1,\ldots, n$}
        \STATE{$S_B \leftarrow B$ uniformly sampled training points}
        \IF{\texttt{DoSGD}$(\mathcal M, B, S_B)$}
            \STATE{$\mathcal M \leftarrow \texttt{IS\_SGD}(\mathcal M, S_B, x)$}
        \ELSE 
            \STATE{$\mathcal M \leftarrow \texttt{Vanilla\_SGD}(\mathcal M, S_B, x)$}
        \ENDIF
    \ENDFOR
    \STATE{$y \leftarrow $ validation error of $\mathcal M$}
    \STATE{$c \leftarrow$ time used to train $\mathcal M$}
    \STATE{$\mathcal D_n \leftarrow \mathcal D_n \cup\{(x,B,y,c)\}$}
\ENDFOR
\STATE \textbf{return} $x^* \in \{x_i\}$ with best predicted error at $B^*$
\end{algorithmic}
\end{algorithm}

\begin{rmk}
Whereas Fabolas speeds up the evaluation of $f(x)$ by limiting the number of training points used during training, \model uses the entire training data, reweighting points based on their relevance to the training task. Thus, each \model iteration is slower than a Fabolas iteration. However, because \model carries out a more principled search through hyperparameter space and queries higher fidelity evaluations, \model requires less BO iterations --- and hence potentially less time --- to find a good hyperparameter.
\end{rmk}

\section{Experiments}
\label{sec:experiments}

\begin{table*}[t]
   \caption{\small Test error of models trained using hyperparameters found by the different methods. Each method is allocated the same amount of time; results reflect each method's choice of best hyperparameter after the different percentages of time have elapsed. Test error is obtained by a model trained on the full training set using vanilla minibatch SGD. Across all three experiments, \model reaches the lowest test error, confirming that the computational cost incurred by importance sampling is amortized by the more efficient search over hyperparamater space that IS enables. Notably, \model also achieves lower test errors earlier than other methods on the more difficult benchmarks.}
   \vspace{-1em}
   \label{tab:summary}
    \begin{center}
    \scalebox{.86}{
        \begin{tabular}{ccccccc}
        \toprule
         \textsc{Problem} & \textsc{Time Budget}                       & IBO (ours)           & \textsc{Fabolas}    & \textsc{Fabolas-IS} & ES                  & ES-IS               \\\midrule
         \multirow{4}{*}{\textsc{CNN (CIFAR10)}} & $\mathit{25}$\%  & \bf 0.28~(0.25,0.32)    & 0.29~(0.26,0.36)    & 0.4~(0.38,0.9)  &  0.38~(0.29,0.83)   & 0.3~(0.26,0.35)\\
                                               & $\mathit{50}$\%  & 0.27~(0.26,0.29)    & 0.26~(0.25.0.27)    & 0.38~(0.27,0.9)   & 0.28~(0.27.0.37)  & \bf 0.25~( 0.25,0.26)  \\
                                               & $\mathit{75}$\%  &\bf  0.25(0.24,0.29)     &  \bf 0.25~(0.24.0.27)   & 0.38~(0.26,0.9)     & 0.28~(0.26.0.28) &  0.26~(0.25,0.26)  \\
                                               & $\mathit{100}$\% & \bf 0.23~(0.23,0.23) &  0.25~(0.24.0.27)  &  0.33~(0.26,0.38)  & 0.28~(0.26.0.28)  &  0.26~(0.25,0.26)   \\\midrule
         \multirow{4}{*}{\textsc{ResNet (CIFAR10)}} & $\mathit{25}$\% 
         & \bf 0.11~(0.11,0.12)    & \bf 0.11~(0.11,0.12)    &\bf  0.11~(0.11,0.2)  & 0.12~(0.11,0.21)    & \bf 0.11~(0.11,0.21) \\
                                               & $\mathit{50}$\%  &\bf 0.1~(0.1,0.1)    & 0.11~(0.1,0.11)     &  0.11~(0.11,0.11)    &  0.11~(0.1,0.2)   &  0.12~(0.11,0.21)   \\
                                               & $\mathit{75}$\%  & \bf 0.09~(0.09,0.1)     &   0.11~(0.1,0.11)    &  0.11~(0.11,0.11)     &  0.11~(0.1,0.17)  &   0.12~(0.11,0.19)   \\
                                               & $\mathit{100}$\% & \bf  0.09~(0.09,0.1)   & 
                                   0.11~(0.1,0.11)   & 
                                   0.11~(0.11,0.11)   & 
                                   0.11~(0.1,0.17)   &  
                                   0.12~(0.11,0.18) )   \\\midrule
         \multirow{4}{*}{\textsc{ResNet (CIFAR100)}} & $\mathit{25}$\%  & 0.38~(0.35,0.39)    & \bf 0.37~(0.37,0.39)   & 0.39~(0.38,0.44)  &  0.38~(0.38,0.44)  & 0.38~(0.37,0.38)\\
                                               & $\mathit{50}$\%  & \bf 0.33~(0.33,0.37)    &0.37~(0.36,0.39)   & 0.39~(0.38,0.44)  &  0.38~(0.38,0.44)  & 0.37~(0.36,0.38)  \\
                                               & $\mathit{75}$\%  & \bf 0.33~(0.33,0.34)    &0.37~(0.36,0.39)   & 0.39~(0.38,0.44)  &  0.38~(0.38,0.42)  & 0.38~(0.36,0.39)  \\
                                               & $\mathit{100}$\% & \bf 0.32~(0.32,0.34)    &0.37~(0.36,0.39)   & 0.39~(0.39,0.45)  &  0.38~(0.37,0.41)  & 0.36~(0.34,0.38)  \\
                                               \bottomrule
        \end{tabular}
    }
    \end{center}
    \vspace{-1em}
\end{table*}

We evaluate our proposed method, \model,\footnote{The code for \model will be released upon acceptance.} on four benchmark hyperparameter tuning tasks: a feed-forward network on MNIST, a convolutional neural network (CNN) on CIFAR-$10$, a residual network on CIFAR-$10$, and a residual network on CIFAR-$100$. We include the following baselines:
\begin{itemize}[label={--},leftmargin=*,parsep=0pt]
    \item \textbf{ES}: Bayesian optimization with the entropy search acquisition function~\citep{hennig2012entropy},

    \item \textbf{ES-IS}: BO with entropy search; inner optimization is performed using IS. For each black-box query, we draw the presample size $B$ uniformly at random from $\{2, \ldots,6\} \times$ batch size as prescribed in~\citep{katharopoulos2018not}; $B$ is constant during the $n$ rounds of SGD. 
    \item \textbf{Fabolas}~\citep{klein2016fast}: BO in which each inner-loop optimization uses a fraction $s$ of the training set. The value of $s$ is learned via multi-task BO; this sub-training set does not evolve during the inner SGD iterations.
    \item \textbf{Fabolas-IS}: Fabolas, training with SGD-IS. For this method, a fraction $s$ of the training data is uniformly chosen as in Fabolas, but training is performed with SGD-ISo. The pre-sample batch size $B$ is the randomly uniformly sampled in $\{2, \ldots,6\} \times$ batch size.

\end{itemize}
ES-IS acts as an ablation test for IBO's multi-task framework, as it does not reason about the cost-fidelity tradeoff of IBO. Thus, we keep the training procedure for ES-IS and Fabolas-IS similar to IBO, switching to IS only if variance reduction is possible and using IS is advantageous (Alg.~\ref{alg:IBO}, lines $8-11$). We run all methods on a PowerEdge R730 Server with NVIDIA Tesla K80 GPUs (experiment \ref{exp:FFNN-MNIST}) or on a DGX-$2$ server with NVIDIA Tesla V100 GPUs (rest).

\subsection{Implementation Details} 
For IBO, we use task kernels $k^{(f)}_B$ (Eq.~\ref{eq:task-kernel}) and $k^{(c)}_B$ (Eq.~\ref{eq:cost-kernel}), with kernel hyperparameters $\lambda=1$ and $\nu=2$. Following \citet{snoek2012practical}, we marginalize out the GPs' hyperparameters using MCMC for all methods. 

To set the time budget, we fix a total number of BO iterations for each method; the time at which the fastest method completes its final iteration acts as the maximum amount of time available to any other method. All initial design evaluations also count towards the runtime; this slightly advantages non-IS methods, which have cheaper initializations. 

We report the performance of each method as a function of wall-clock time, since the methods differ in per-iteration complexity (App.~\ref{fig:inc-diff} reports results vs. iteration number). 

We measure the performance of each method by taking the predicted best hyperparameter values $x^*$ after each BO iteration, then training a model with hyperparameters $x^*$, using the \emph{entire} training set and vanilla SGD. Recall that for Fabolas, Fabolas-IS, and IBO, the incumbent $x^*$ is the set of hyperparameters with the best predicted objective on the target task (\eg, using the full training data for Fabolas).

We run each method five times unless otherwise stated, and report the median performance and $25^{th}$ and $75^{th}$ percentiles (mean and standard deviation results are included in Appendix \ref{app:mean-results} for completeness). ES and Fabolas variations are run using RoBO.\footnote{\url{https://github.com/automl/RoBO}} For importance sampling, we used the code provided by~\citet{katharopoulos2018not}.\footnote{\url{https://github.com/idiap/importance-sampling}} 

All methods are initialized with 5 hyperparameter configurations drawn from a Latin hypercube design. For IBO, we evaluate each configuration on the maximum value of its target task $B$. For Fabolas, ~\citet{klein2016fast} suggest initializing by evaluating each hyperparameter on an increasing series of task values. This aims to capture the task variable's effect on the objective. However, we empirically observed that following an initial design strategy similar to IBO's, \ie, evaluating each hyperparameter on the maximum target value $s$, worked better in practice for both Fabolas and Fabolas-IS. This is the method we use in our experiments;  App.~\ref{app:fab-init-comparison} includes results for both initialization schemes.

For \model, Fabolas-IS and ES-IS, we reparameterize the pre-sample size $B$ as $B = b \times s_B$. As was recommended by \citet{katharopoulos2018not}, we set $s_B \in [2, 6]$. For Fabolas-IS, if $B$ is larger than the training subset size, we use the entire subset to compute the importance distribution.
\subsection{Feed-forward Neural Network on MNIST}\label{exp:FFNN-MNIST}%
\vspace{-.5em}
Our first experiment is based on a common Bayesian optimization benchmark problem~\citep{falkner2018bohb,domhan2015speeding,hernandez2016predictive}. We tune a fully connected neural network using RMSProp on MNIST~\citep{lecun1998mnist}. The number of training epochs $n$ and the number of BO rounds are set to $50$. We tune six hyperparameters: number of hidden layers, number of units per layer, batch size, learning rate, decay rate, and dropout rate (see App.~\ref{app:mnist}). 
\begin{figure*}[h!]
\begin{minipage}[t]{0.48\textwidth}
\includegraphics[width=\textwidth]{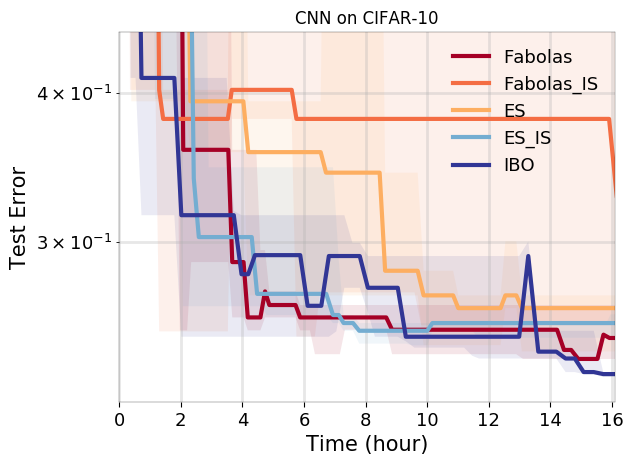}
\vspace{-2em}
\caption{\small Hyper-parameter tuning of a CNN on CIFAR-10. After around 9 hours, our model (\model) outperforms all other methods. The ablation test Fabolas-IS shows the weakest performance with a large uncertainty. The ablation test ES-IS shows slightly better performance than \model in the first half of the time horizon. However, \model overall surpasses ES-IS and achieves the best final performance among all methods with a negligible variance, confirming the value of our multi-task formulation.}
\label{fig:cnn-cifar10-wall}
\end{minipage}\hfill
\begin{minipage}[t]{0.48\textwidth}
\includegraphics[width=\textwidth]{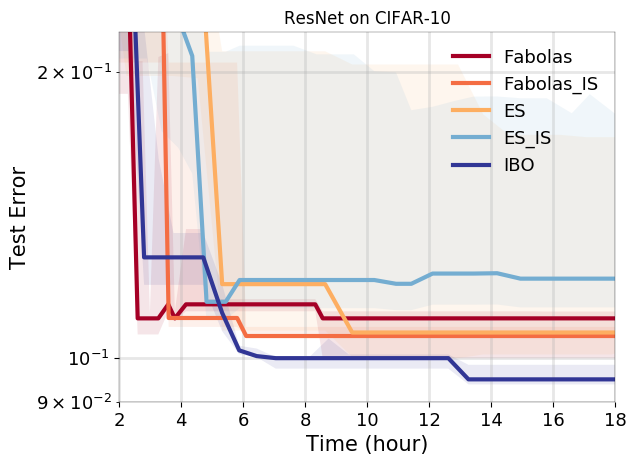}
\vspace{-2em}
\caption{\small Hyper-parameter tuning of a ResNet on CIFAR-10. \model outperforms all other baselines at one third of the time budget and keep improving until the end. Conversely, Fabolas-IS  is unable to progress after one third of the time horizon while Fabolas achieves a minor improvement (compared to \model) at around 9 hours. ES-IS shows the weakest performance of all and  suffers a large uncertainty. This is yet another evidence that simply augmenting BO with importance sampling is not robust.}
\label{fig:resnet-cifar10-wall}
\end{minipage}
\vspace{-1em}
\end{figure*}

Given the well-known straightforwardness of the MNIST dataset, we do not expect to see significant gains when using importance sampling during training. Indeed, we see (Table~\ref{tab:summary-supp}) that all methods perform similarly after exhausting their BO iteration budget, although Fabolas does reach a low test error slightly earlier on, since training on few data points is sufficient; see Appendix~\ref{app:mnist} for more details.

\subsection{CNN on CIFAR-10}
\label{exp:CNN-CIFAR10}
\vspace{-.5em}
We next tune a convolutional neural network (CNN) using RMSProp on the CIFAR-10 dataset~\citep{krizhevsky2009learning}. We fix an architecture of three convolutional layers with max-pooling, followed by a fully connected layer, in line with previous benchmarks on this problem~\citep{falkner2018bohb,klein2016fast,dai2019bayesian}. Following~\citet{dai2019bayesian}, we tune six hyperparameters: number of convolutional filters $n_c \in \{128,\dots,256\}$, number of units in the fully connected layer $n_u \in\{256,\dots,512\}$, batch size $b \in \{32,\dots,512\}$, initial learning rate $\eta \in [10^{-7},0.1]$, decay rate $\beta \in[10^{-7},10^{-3}]$, and regularization weight $ \upsilon \in[10^{-7},10^{-3}]$.
All methods are run for $100$ BO iterations and trained using $n=50$ SGD epochs. 

IBO, Fabolas and ES-IS exhibit the best performance (Fig.~\ref{fig:cnn-cifar10-wall}) but switch ranking over the course of time. However, after spending roughly half of the budget, IBO outperforms Fabolas and all other baselines, achieving the best final error with the lowest uncertainty.

ES-IS shows that adding IS naively can improve upon base entropy search; however, IBO outperforms both ES and ES-IS, confirming the importance of a multi-task setting that optimizes IS. Furthermore, simply adding importance sampling during SGD is not guaranteed to improve upon any method: Fabolas-IS performs poorly compared to Fabolas.
\vspace{-.5em}
\subsection{Residual Network on CIFAR-10}
\label{exp:ResNet-CIFAR10}
\vspace{-.5em}
We next tune the a residual network trained on CIFAR-10. We follow the wide ResNet architecture in \citep{zagoruyko2016wide}, and tune four hyperparameters: initial learning rate $\eta \in [10^{-6},~1]$, decay rate $\beta \in [10^{-4},~1]$, momentum $\omega \in [0.1,0.999]$ and $L_2$ regularization weight $ \upsilon \in[10^{-6},~1]$. Following \cite{klein2016fast}, all but the momentum are optimized over a log-scale search space. 

\begin{figure*}[t!]
\begin{minipage}[t]{0.48\textwidth}
\includegraphics[width=\textwidth]{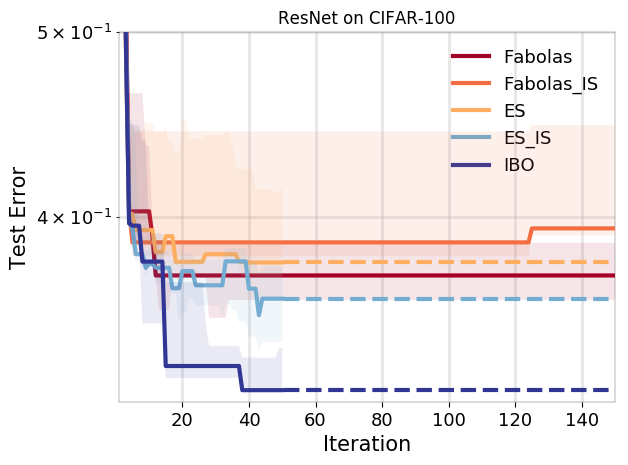}
\end{minipage}\hfill
\begin{minipage}[t]{0.48\textwidth}
\includegraphics[width=\textwidth]{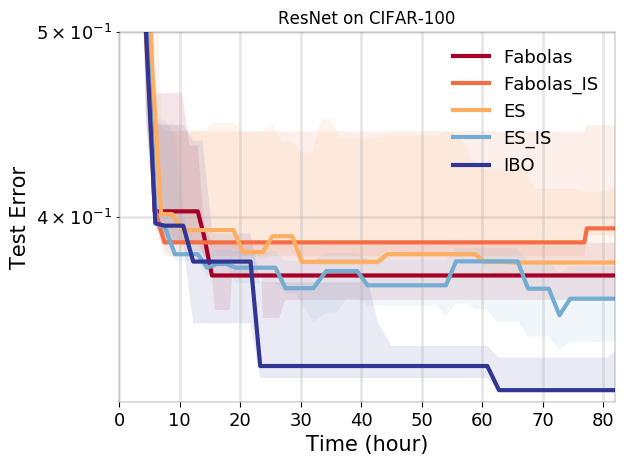}
\label{fig:screwthread}
\end{minipage}
\vspace{-1.5em}
\caption{\small Hyper-parameter tuning of a ResNet on CIFAR-100.  \model outperforms all other methods as both a function of iterations and time. The performance over iteration (left plot) in particular shows that \model is able to achieve a low test error in a limited number of function evaluations, i.e., less than 20. This roughly equals spending 20 $\%$ of the time budget (right plot). Moreover, \model is able to further improve after 60 hours. Conversely, both Fabolas and Fabolas-IS are unable to progress after an early stage (roughly 10 out of 150 iterations and 15 out of 80 hours). Interestingly, providing a larger iteration budget has not helped these methods. ES-IS exhibits the second best performance after \model with 4-5 $\%$ margin. Moreover, compared to \model, it shows a noisier performance with larger uncertainty. }
\label{fig:resnet-cifar100-iter-time}
\end{figure*}

We set $n=50$ and multiply the learning rate by the decay rate after $n=40$ epochs. Experimentally, we saw that $n=50$ epochs is insufficient for the inner (SGD) optimization to converge on the ResNet architecture; this experiment evaluates BO in the setting where $f$ is too computationally intensive to compute exactly. We ran all the methods using $80$ BO iterations for Fabolas and Fabolas-IS and $50$ iterations for the rest. This difference in budget iteration is to compensate for the different cost of training on a subset of data versus on the entire data.\footnote{For experiment~\ref{exp:FFNN-MNIST}, we observed that keeping the BO iteration budget consistent is sufficient since the training costs are not very different.  For experiment~\ref{exp:CNN-CIFAR10}, we set this budget to 100, and stopped reporting the results once the first method exhausted its budget. Since ResNet experiments are generally more costly, choosing a large budget for all methods was not feasible.} Results are reported in Fig.~\ref{fig:resnet-cifar10-wall}, and are obtained with $3$ runs with random initializations.

Consistently with previous results, Fabolas achieves the lowest error in the very initial stage, due to its cheap approximations. However, \model quickly overtakes all other baselines, and attains a value that other methods cannot achieve with their entire budget consumption. Fabolas-IS also performs well, but suffers a large variance.

The ablation tests (ES-IS and Fabolas-IS) consistently have high variance, likely because these methods do not learn the optimal batch size for importance sampling and opt for a random selection within the recommended range. In contrast, \model specifically learns the batch size parameter which controls the cost-benefit trade off in importance sampling and hence, enjoys better final results and lower variance.

\subsection{Residual Network on CIFAR-100}
Finally, we tune the hyperparameters of a residual network trained on CIFAR-100. The architecture of the network, the hyperparameters we optimize and their respective ranges are similar to the \S\ref{exp:ResNet-CIFAR10}. We set $n=200$ and multiply the learning rate by the decay rate every $40$ epochs. For Fabolas and Fabolas-IS, a budget of $150$ BO iterations is provided while the rest of the methods are given $50$ iterations. 

Clearly, \model outperforms the rest of the methods after spending roughly 20 $\%$ of the time budget (Fig.~\ref{fig:resnet-cifar100-iter-time}); Fabolas and ES-IS are the second best methods. Similar to the experiment~\ref{exp:CNN-CIFAR10}, Fabolas-IS is outperformed by the other baselines, and once again incurs a large variance. Interestingly, for Fabolas and Fabolas-IS, the additional BO budget does not cause an improvement in their performance. This is yet further evidence that for complex datasets, neither vanilla multi-task frameworks nor simple importance sampling is sufficient to gain the advantages of \model.

By seeking higher-fidelity surrogate models, \model achieves better results in fewer optimization runs and less runtime than other baselines, despite the incurred cost of using each training example individually during certain SGD rounds.

\section{Conclusion}
Bayesian optimization offers an efficient and principled framework for hyperparameter tuning. However, finding optimal hyperparameters requires an expensive inner loop which repeatedly trains a model with new hyperparameters. Prior work has scaled BO by using cheap evaluations to the black-box function. \model takes the opposite approach: by increasing time spent obtaining higher-fidelity evaluations, \model requires much fewer outer BO loops. 

Leveraging recent developments in importance sampling, \model takes into account the contribution of each training point to decide whether to run vanilla SGD or a more complex, time-consuming but higher quality variant. Although this results in costlier neural network training loops, the additional precision obtained for the black-box estimates allows a more principled search through hyperparameter space, significantly decreasing the amount of wall-clock time necessary to obtain a high-quality hyperparameter. 

Crucially, the interaction between importance sampling and Bayesian optimization must be approached with care; a naive merging of both methods does not decrease the overall runtime of Bayesian optimization, and does not yield better final hyperparameters. However, by opting for a multi-task parameterization of the problem, \model learns to dynamically adjust the trade-off between neural network training time and black-box estimate value, producing faster overall runtimes \emph{as well as} better hyperparameters. 

We show on four benchmark tasks of increasing complexity that \model achieves the lowest error compared to all other baseline methods, and scales gracefully with dataset and neural architecture complexity. When tuning a ResNet on CIFAR-100, \model outperforms all other baselines and ablation tests by a significant margin, both as a function of wall-clock time and number of outer optimization rounds.

\bibliography{references}
\bibliographystyle{authordate1}
\clearpage
\appendix
\section{Feed-forward Neural Network on MNIST}
\label{app:mnist}
\begin{figure}[ht!]
  \hspace*{0.4cm}   \includegraphics[width=0.72\linewidth]{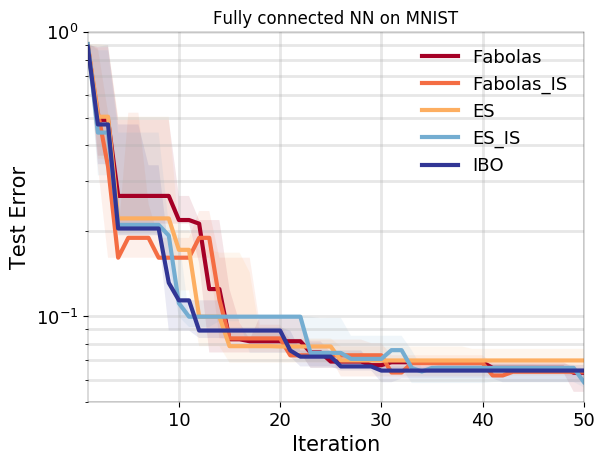}\par 
  \hspace*{0.4cm} 
    \includegraphics[width=0.72\linewidth]{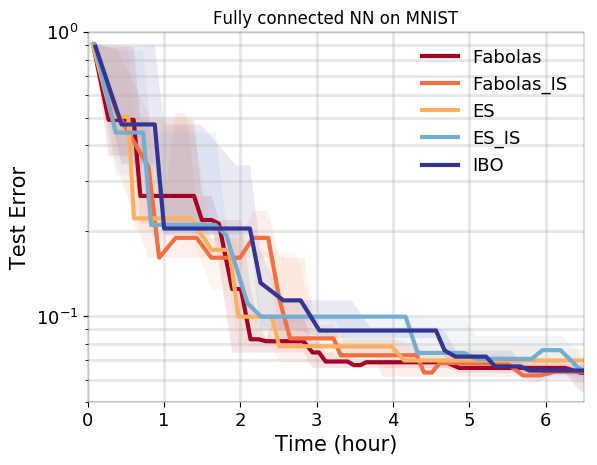}\par 
\caption{Average performance of all methods on MNIST as a function of both iteration budget (left column) and wall-clock time (right column).}
\label{fig:mnist}
\end{figure}

Per BO iteration (Fig. \ref{fig:mnist}), IBO is amongst the best performing methods gaining higher utility compared to the others. However, since performing importance sampling is expensive, IBO's performance  degrades over wall-clock time. After spending roughly 30$\%$  of the time budget (around two hours), Fabolas outperforms the other methods. This is expected since Fabolas utilizes cheap approximations by using training subsets. Although such approximations are noisy, we speculate that it does not significantly harm the performance, specially for simpler datasets and models such as a feed-forward network on MNIST.

We tune six hyperparameters: number of hidden layers $n_\ell \in \{1,\dots,5\}$, number of units per layer $n_u \in\{16,\dots,256\}$, batch size $b \in \{8,\dots,256\}$, initial learning rate $\eta \in [10^{-7},\dots,~10^{-1}]$, decay rate $\beta \in [10^{-7},~10^{-3}]$ and dropout rate $\rho \in [0,~0.5]$. Following~\citep{falkner2018bohb}, the batch size, number of units, and learning rate are optimized over a log-scale search space.

All methods are run for $50$ BO iterations. The performance is averaged over five random runs and shown in the last row of Figure \ref{fig:inc-diff} (median with 25 and 75 percentiles over time and iteration budget) and Figure \ref{fig:mean-res} (mean with standard deviation over time).
\begin{table}[ht!]
\caption{Test error of models trained using hyperparameters found by the different methods. Each method is allocated the same amount of time; results reflect each method's choice of best hyperparameter after the different percentages of time has elapsed. Test error is obtained by a model trained on the full training set using vanilla minibatch SGD. All methods roughly perform similar achieving 6 $\%$ error at max budget. However, Fabolas starts its progress earlier. Given the simplicity of MNIST, we speculate that the cheap noisy approximations provided by Fabolas via uniform sampling, suffices to attain improvement while importance sampling is unnecessarily costly.}
\label{sample-table}
\vskip 0.15in
\begin{center}
\begin{small}
\begin{sc}
\begin{tabular}{lcccccr}
\toprule
Metric & Method  &$\mathit{25}$\%  & $\mathit{50}$\% & $\mathit{75}$\% & $\mathit{100}$\% \\
\midrule
&IBO& 0.2 &0.09&0.07&\bf 0.06    \\
&Fabolas&0.21 &\bf0.07&\bf0.06&\bf0.06   \\
Median &Fabolas-IS &\bf0.16 &\bf0.07&0.07&\bf0.06   \\
Error&ES     & 0.17&0.08&0.07&0.07        \\
&ES-IS &0.21 &0.1&0.07&\bf0.06     \\

\bottomrule
\end{tabular}
\end{sc}
\begin{sc}
\begin{tabular}{lcccccr}
\toprule
Metric & Method  &$\mathit{25}$\%  & $\mathit{50}$\% & $\mathit{75}$\% & $\mathit{100}$\% \\
\midrule
&IBO& 0.19&0.08&0.07&0.06    \\
&Fabolas& 0.15&0.07&0.06&0.05   \\
25 $\%$ &Fabolas-IS  & 0.16&0.07&0.06&0.06   \\
Error&ES     &0.12 &0.07&0.07&0.06        \\
&ES-IS &0.19 &0.1&0.07&0.06     \\

\bottomrule
\end{tabular}
\end{sc}
\begin{sc}
\begin{tabular}{lcccccr}
\toprule
Metric & Method  &$\mathit{25}$\%  & $\mathit{50}$\% & $\mathit{75}$\% & $\mathit{100}$\% \\
\midrule
&IBO& 0.34&0.11&0.08&0.06    \\
&Fabolas&0.22 &0.07&0.06&0.06   \\
75 $\%$ &Fabolas-IS  & 0.19&0.08&0.07&0.06   \\
Error&ES     & 0.2&0.08&0.08&0.08        \\
&ES-IS & 0.21&0.1&0.07&0.06     \\

\bottomrule
\end{tabular}
\end{sc}
\end{small}
\end{center}
\vskip -0.1in
\label{tab:summary-supp}
\end{table}

\section{IBO scales with dataset and network complexity}
\begin{figure*}
\begin{multicols}{2}
    \includegraphics[width=0.75\linewidth]{figures/ResNet_CIFAR100/plot_average_results_over_iter_with_median_log_scale_MTBO_choice_target_task.png}\par 
    \includegraphics[width=0.75\linewidth]{figures/ResNet_CIFAR100/plot_average_results_over_time_with_median_log_scale_MTBO_choice_target_task.png}\par 
\end{multicols}
\begin{multicols}{2}
    \includegraphics[width=0.75\linewidth]{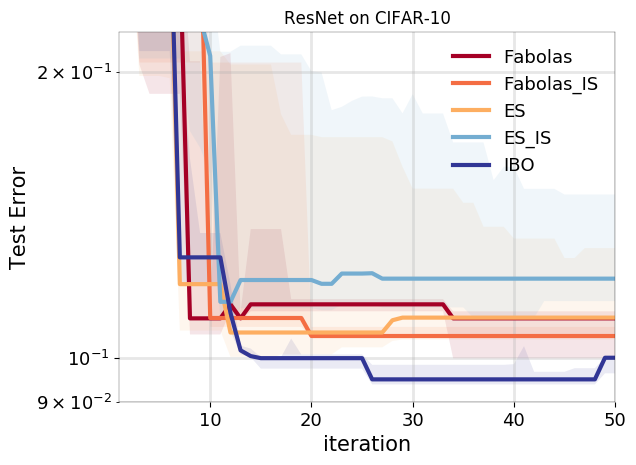}\par 
    \includegraphics[width=0.75\linewidth]{figures/ResNet_CIFAR10/plot_average_results_over_time_with_median_log_scale_MTBO_choice_target_task.png}\par 
\end{multicols}
\begin{multicols}{2}
    \includegraphics[width=0.75\linewidth]{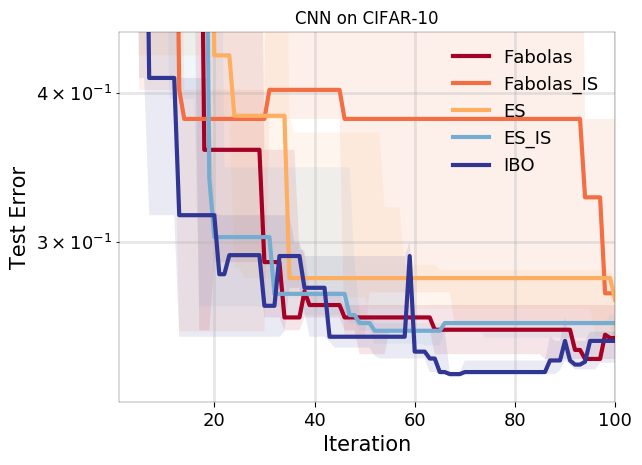}\par 
    \includegraphics[width=0.75\linewidth]{figures/CNN_CIFAR10/plot_average_results_over_time_with_median_log_scale_MTBO_choice_target_task.png}\par 
\end{multicols}
\begin{multicols}{2}
  \hspace*{0.4cm}   \includegraphics[width=0.72\linewidth]{figures/FFNN_MNIST/plot_average_results_over_iter_with_median_log_scale_MTBO_choice_target_task.png}\par 
  \hspace*{0.4cm} 
    \includegraphics[width=0.72\linewidth]{figures/FFNN_MNIST/plot_average_results_over_time_with_median_log_scale_MTBO_choice_target_task.png}\par 
\end{multicols}
\caption{Average performance of all methods for all experiments as a function of both iteration budget (left column) and wall-clock time (right column). Each row represents one experiment such that the difficulty of tuning increases from the bottom row to the top
 i.e., the most straight-forward problem, FCN on MNIST, lies in the bottom row and the most challenging benchmark, ResNet on CIFAR100 is in the top row. In the iteration plots, IBO is consistently amongst the best methods (lower curve denotes better performance), achieving high utility per BO iteration. However, since doing importance sampling is inherently expensive, the advantage of IBO over wall-clock time gradually manifests once the tuning becomes more challenging. Specifically,  IBO starts to outperform the rest earlier with an increasing margin over wall-clock time, the more difficult benchmarks become (from the bottom row to the top).}
\label{fig:inc-diff}
\end{figure*}

IBO improves upon existing BO methods, moreso when tuning on large complex datasets and architectures. To illustrate, Figure \ref{fig:inc-diff} includes the results of all experiments over iteration budget (left column) and wall-clock time budget (right column). Moreover, the plots are sorted such that complexity of dataset and model architecture decreases along the rows; i.e., the most straight-forward problem, FCN on MNIST, lies in the bottom row and the most challenging experiment, ResNet on CIFAR100 is in the top row. In the iteration plots (left column), IBO is consistently amongst the best methods (lower curve denotes better performance), achieving high utility per BO iteration. However, since doing importance sampling is inherently expensive, the advantage of IBO over wall-clock time gradually manifests once the tuning becomes more challenging. Specifically, moving from the bottom to the top, as the complexity level of tuning increases, IBO starts to outperform the rest from and earlier stage and with an increasing margin over wall-clock time (right column).

\section{Initializing Fabolas}
\label{app:fab-init-comparison}
\begin{figure}[H]
\begin{multicols}{2}
    \includegraphics[width=1\linewidth]{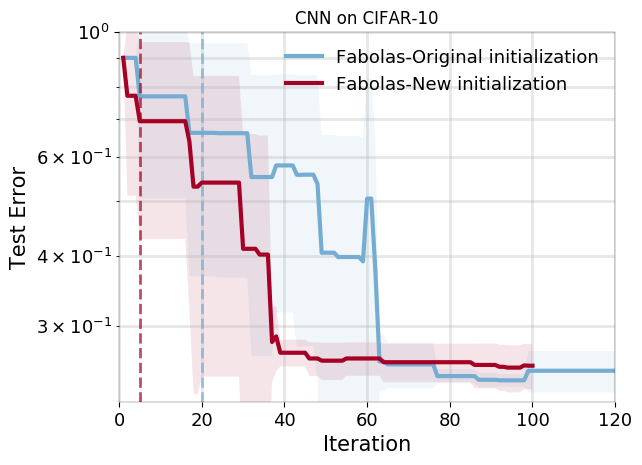}\par
    \includegraphics[width=1\linewidth]{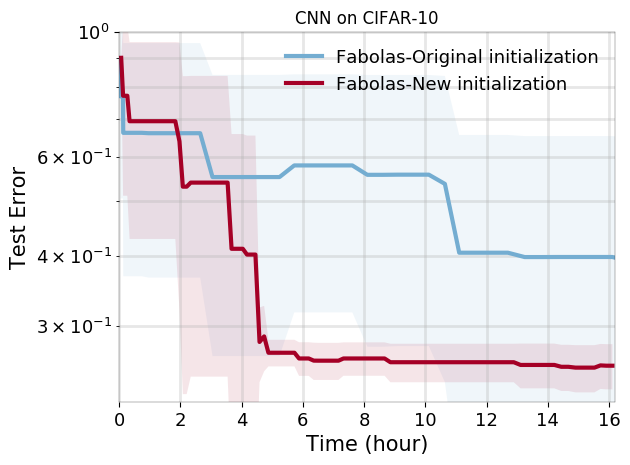}\par
\end{multicols}

\begin{multicols}{2}
    \includegraphics[width=1\linewidth]{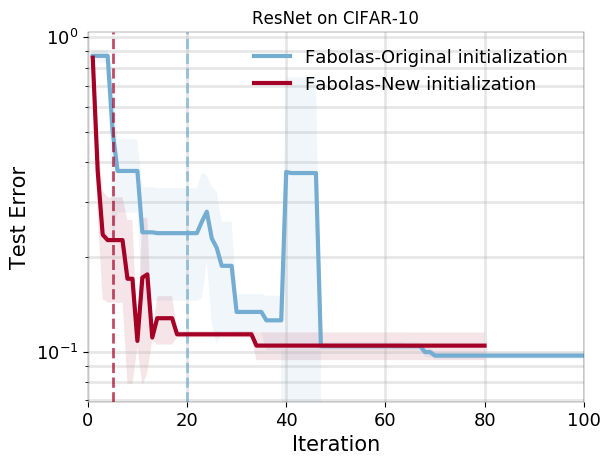}\par
    \includegraphics[width=1\linewidth]{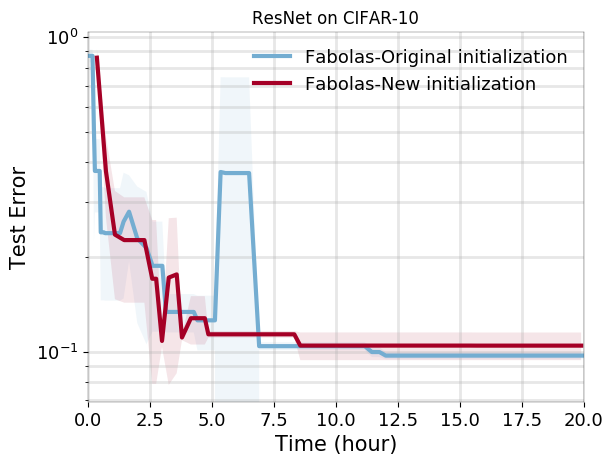}\par
    \end{multicols}

\caption{Comparison between two initialization schemes of Fabolas for CNN and ResNet on CIFAR-10. The dashed lines (left column) show the number of initial design evaluations for each method, immediately followed by the start of BO. We observe that with the new initial design scheme, Fabolas can potentially start progressing at a smaller iteration and a lower time, and achieve a reduced  variance.}
\label{fig:fabolas-init}
\end{figure}

Conventionally, Bayesian optimization starts with evaluating the objective at an initial set of hyperparameters chosen at random. To leverage speedup in Fabolas, \citet{klein2016fast} suggests to evaluate the initial hyperparameters at different, usually small, subsets of the training data. In our experiments, we randomly selected $5$ hyperparameters and evaluated each on randomly selected training subsets with sizes $\{\frac{1}{128},\frac{1}{64},\frac{1}{32},\frac{1}{16}\}$ of the entire training data. However, our experimental results show that Fabolas achieves better results faster if during the initial design phase, the objective evaluation use the entire training data. 
Figure \ref{fig:fabolas-init} illustrates this point for CNN and ResNet on CIFAR-10. Fabolas with the original initialization scheme performs $20$ evaluations ($5$ hyperparameters each evaluated at $4$ budgets) where with the new scheme, Fabolas initializes with $5$ evaluations ($5$ hyperparameters each evaluated at $1$ budget). The plots show the mean results (with standard deviation) averaged over five and three runs for CNN and ResNet. Overall, the Fabolas with new initialization achieves better average performance.
\begin{figure}[ht!]
  \hspace*{0.4cm}   \includegraphics[width=0.72\linewidth]{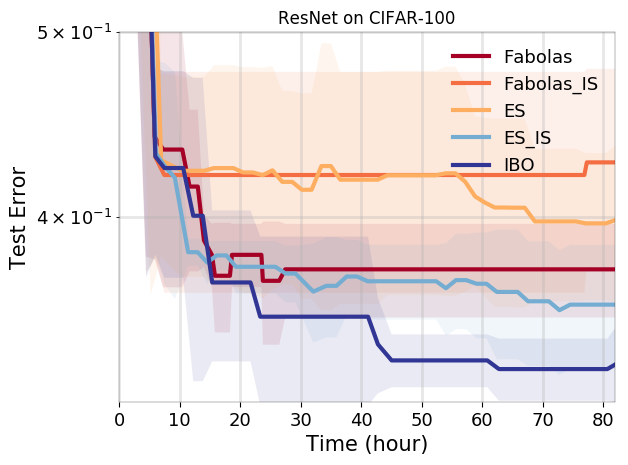}
  \par 
  \hspace*{0.4cm} 
  \includegraphics[width=0.72\linewidth]{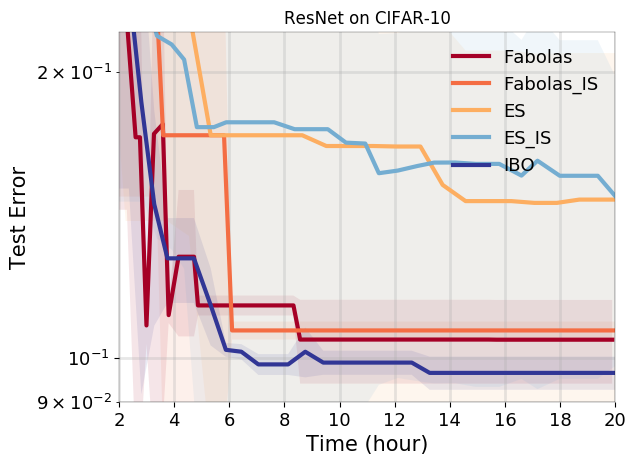}\par 
  \hspace*{0.4cm} 
\includegraphics[width=0.72\linewidth]{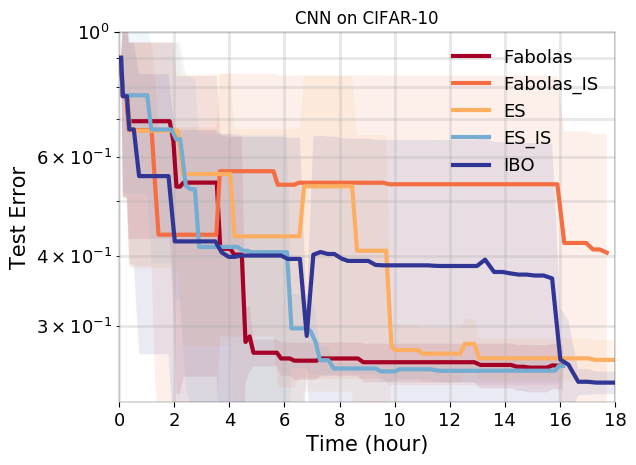}\par 
  \hspace*{0.4cm} 
\includegraphics[width=0.72\linewidth]{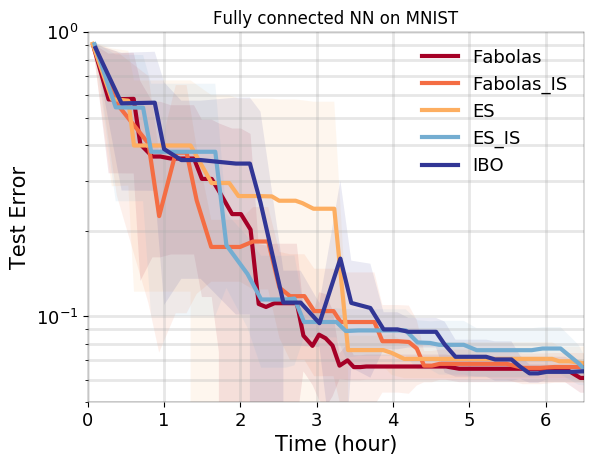}\par 

\caption{These plots show the mean performance (with standard deviation) of all methods for all the experiments. \model consistently achieves amongst the lowest test errors at the maximum budget. For CNN on CIFAR10, \model suffers 1 relatively weak run (out of 5 total runs) which affects the mean and standard deviation . For a different perspective, see Fig.~\ref{fig:inc-diff} reporting median and 25/75 $\%$. }
\label{fig:mean-res}
\end{figure}

\subsection{Mean and Standard Deviation Results}
\label{app:mean-results}
For completion, we include the plots reporting mean and standard deviation throughout the experiments (Figure \ref{fig:mean-res}).

\end{document}